# Physics-informed generative AI for semiconductor manufacturing: *Enforcing hard physical constraints in generative models by construction*

Yaser Mike Banad, [1,3,4*] Sarah Sharif [1,2,3,4]

[1]School of Electrical and Computer Engineering, University of Oklahoma

[2]Center for Quantum Research and Technology, University of Oklahoma

[3]Intelligent Neuromorphic and Quantum Understanding for Innovative Research and Engineering (INQUIRE) Laboratory

[4]Material Science and Engineering Program, University of Oklahoma, Norman, OK 73019 USA

[*] Corresponding Author, email: bana@ou.edu

**Abstract.**

Generative models are increasingly used to propose designs, data, and control actions for physical systems, yet many such systems are governed by hard physical constraints rather than by perceptual plausibility. Semiconductor manufacturing provides a demanding test case: generated masks, layouts, synthetic defect data, and process recipes must obey lithography, transport, reaction, and device-physics constraints, because physically invalid samples are not merely low quality but unusable. This Perspective argues that semiconductor manufacturing exposes a broader computational-science challenge, namely that generative AI for constrained physical domains must be physics-informed by construction, not corrected only through post-hoc filtering. We survey the emerging architectural toolkit, including physics-informed diffusion, PDE-constrained variational models, neural-operator priors, and conservation-law-respecting generative networks, and show how it connects to differentiable lithography, TCAD, process simulation, and autonomous experimentation. We identify four integration patterns between generative models and physics-based simulators, and we propose a research agenda centered on physics-fidelity benchmarks, differentiable simulator infrastructure, and multimodal foundation models for physical design and manufacturing. The central claim is analytical rather than rhetorical: where physical validity is the binding criterion of success, architectures that enforce it by construction should be expected to outperform those that filter for it after the fact, and the fab is the setting where this distinction is sharpest.

## Introduction

Generative AI is moving out of the perceptual domains where it first succeeded and into physical ones. Models that began by producing text and images are now asked to propose molecules, materials, device geometries, experimental protocols, and manufacturing recipes. In these settings the criterion of a good sample changes character. A generated image is judged by whether it looks right to a human observer, and an output that is approximately right is often useful. A generated physical artifact is judged by whether it obeys the governing equations of its domain, and an output that violates those equations is frequently not approximately useful but simply unusable. This is a general problem for computational science: how to build generative models for domains in which constraint violation makes an output worthless rather than merely lower in quality. Semiconductor manufacturing is the setting in which this problem is at its most acute, and it therefore serves as a revealing model system for the broader class.

A single line on a process recipe can be the difference between a wafer that yields and a wafer that scraps. Semiconductor manufacturing has always operated under this kind of brittleness, where nanometer-scale deviations from physical law propagate into expensive, sometimes unrecoverable, yield loss. The fab has spent two decades building digital infrastructure (in-line metrology, fault detection and classification, advanced process control, digital twins, MES integration) precisely to keep production inside the narrow process window where physics cooperates with intent. [1], [2], [3] Data-driven machine learning has been layered onto this infrastructure across manufacturing, from deep convolutional defect and quality monitoring to closed-loop process control, [4]–[6] and most recently to hybrid physics-and-data digital twins for advanced semiconductor packaging. [7] Within semiconductor manufacturing specifically, run-to-run and advanced process control have long closed the loop between metrology and recipe adjustment to hold each process on target. [3], [8] This is the same closed-loop, reward-driven paradigm now reshaping autonomous experimentation and self-driving laboratories in adjacent physical-science domains, where Bayesian-optimization and reinforcement-learning agents iterate over surrogate and digital-twin models under human-in-the-loop supervision. [35], [36], [37], [38], [42] Advanced packaging and heterogeneous integration are emerging as a frontier of equal stakes, where warpage, interconnect reliability, and electro-thermomechanical co-design impose physical constraints as binding as the lithographic ones upstream. [9] Into this carefully calibrated stack, generative AI has arrived with extraordinary momentum and limited structural encoding of physical law.

The dominant generative paradigm imported into the fab today is the one that succeeded in language and images: large models trained on internet-scale corpora, sampled to produce plausible outputs, refined through human feedback. [10], [11] This paradigm has produced striking demonstrations in layout and mask generation, process-recipe copilots, synthetic wafer-defect images for inspection, and inverse design of devices. [12]–[17] The demonstrations are real. This trajectory, we argue, is incomplete for production-grade physical systems.

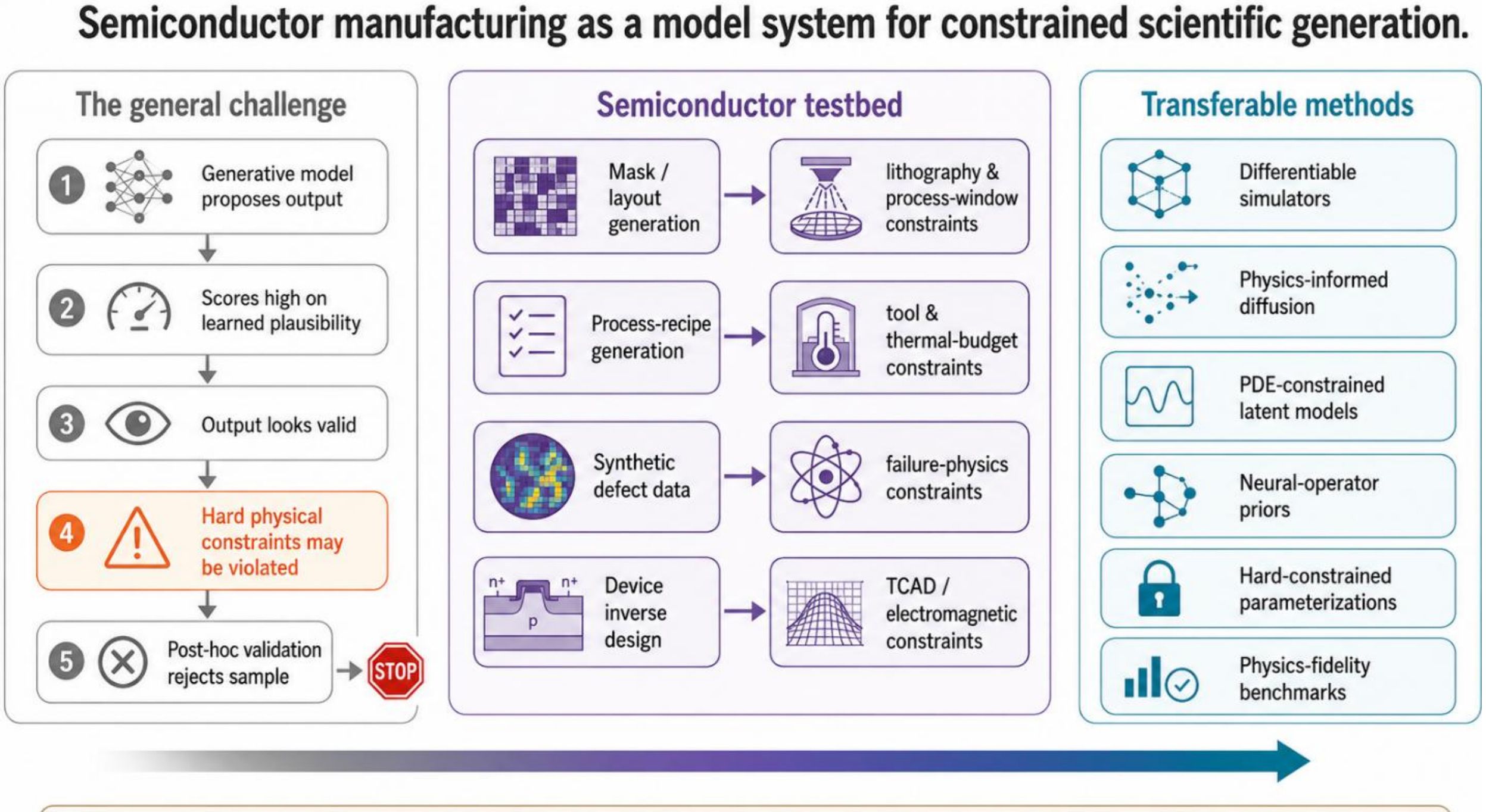

**Figure 1.** Semiconductor manufacturing as a model system for constrained scientific generation. The general computational-science challenge (left): a generative model proposes an output that scores well on learned plausibility yet may violate hard physical constraints, so post-hoc validation rejects it. The semiconductor testbed (center): mask and layout generation, process-recipe generation, synthetic defect data, and device inverse design each face binding lithography, process-window, failure-physics, and TCAD or electromagnetic constraints. The transferable methods (right): differentiable simulators, physics-informed diffusion, PDE-constrained latent models, neural-operator priors, hard-constrained parameterizations, and physics-fidelity benchmarks. The fab is not only an application domain; it is a stress test for generative AI under hard physical constraints.

Semiconductor manufacturing is not a perceptual domain. A generated image that is ninety-five percent photorealistic is useful; a generated mask or recipe that is ninety-five percent physically valid is scrap. The metrics that govern fab success (yield, critical-dimension uniformity, overlay, edge-placement error, defect density, parametric drift) are not approximations of human preference. They are consequences of electromagnetics, transport and reaction physics, and device constitutive behavior. A generative model that does not encode these in its structure will, with high probability, produce outputs that look correct and fail anyway. This failure mode is already visible in early evidence and in well-understood risk. Without lithography-aware constraints, learned generative priors can produce mask and layout candidates that appear plausible under the training distribution yet require independent checks for printability, process-window compliance, and design-rule validity. [14], [15] Language-model process-recipe assistants can likewise propose steps that the qualified toolset cannot execute, and that therefore demand verification against tool and process-window constraints. [12] Synthetic wafer-defect datasets used to train inspection classifiers can drift away from the failure physics they were meant to represent and degrade detection when the underlying process shifts [16].

In this perspective, we argue that the next generation of generative AI for semiconductor manufacturing must be physics-informed by construction, not by post-hoc filtering. The architectural choices made now (what is differentiated, what is conserved, what is sampled, what is constrained) will determine whether generative AI becomes a load-bearing component of the fab stack or remains a productivity gadget restricted to early-stage ideation. We survey the emerging toolkit of physics-informed generative models, identify four concrete integration patterns between the data-driven and physics-based cultures of the fab, and lay out a research agenda centered on differentiable lithography and TCAD simulators and physics-fidelity benchmarks. Our central claim is that, for semiconductor manufacturing, physical validity functions less as a constraint on generative AI than as the operative criterion of success, and that this distinction should drive the architectural choices made now. Figure 1 situates the argument, presenting semiconductor manufacturing as a model system for constrained scientific generation: a general computational-science challenge on the left, the semiconductor testbed in the center, and the transferable methods that address it on the right.

## Semiconductor manufacturing as a model system for constrained generation

Why treat the fab as a model system rather than as one application among many? Because semiconductor manufacturing combines, in a single domain, the features that make constrained generation hard wherever it appears, and it does so with unusual severity (Table 1). The constraints are hard rather than soft: a layout that violates a design rule or a recipe that exceeds a thermal budget cannot be accepted at any preference score, so the perceptual objectives that organize generative modeling elsewhere do not transfer. Validation is expensive: confirming that a candidate is feasible requires lithography, electromagnetic, transport, or full TCAD simulation, or physical qualification on a tool, none of which is cheap enough to run on every sample a generator proposes. The data are sparse, proprietary, and subject to distribution shift: qualified process data are scarce compared with natural images, are rarely shared across organizations, and move whenever a consumable, a chamber, or a tool changes, so pure scaling cannot substitute for structure. The representations are irreducibly multimodal: a single design decision touches natural-language specifications, layout and netlist geometry, inspection imagery, metrology traces, simulated fields, and process recipes, and a useful model must link them. And the decision loop is closed: the fab already operates run-to-run control between metrology and recipe, so generative proposals enter a setting that naturally connects to autonomous experimentation rather than to one-shot sampling.

No single one of these features is unique to the fab. Molecular and materials generation also face hard constraints and expensive validation; robotics and protocol design also close the loop through experiment; scientific imaging is also multimodal. What distinguishes semiconductor manufacturing is that all of these hold at once, at production scale, under economic stakes that make the consequences of an invalid sample immediate and measurable. The methods that succeed here, and the benchmarks that certify them, should therefore transfer outward to the broader class of constrained physical-generation problems, which is why the architectural argument that follows is framed in terms general enough to travel and then grounded in the specifics of the fab.

**Table 1.** Features that make semiconductor manufacturing a model system for constrained scientific generation, and why each matters computationally.

| Feature | Why it matters computationally |
|---|---|
| Hard physical constraints | Invalid generated outputs cannot be redeemed by preference scoring; feasibility is binary, not graded. |
| Expensive validation | Lithography, electromagnetic, transport, and TCAD simulation are costly but necessary, so checking every sample after the fact does not scale. |
| Sparse, proprietary, shifting data | High-quality fab data are scarce and distribution-shifting, so pure scaling is limited and structural priors carry the load. |
| Multimodal representation | Text recipes, layouts, inspection images, metrology traces, and simulated fields must be linked within one model. |
| Closed-loop decision making | Existing run-to-run control connects generation directly to autonomous experimentation rather than to one-shot sampling. |

## Why unconstrained generation fails in physical manufacturing systems

The case for physics-informed architecture rests on three failure modes that are already documented in the semiconductor literature and that are not artifacts of immature models. They are structural consequences of training a generative system to maximize a likelihood objective over historical data without enforcing the physical laws that historical data merely sampled from.

### *Plausible outputs that violate hard constraints*

Generative layout and mask synthesis systems trained on GDSII repositories can produce geometries that are locally smooth and globally unmanufacturable. A diffusion model trained on standard-cell or mask patterns can produce a plausible-looking layout whose features fall below the minimum printable dimension, whose pitches land in the forbidden region of the process, or whose density and assist-feature placement can violate the optical-proximity and design-rule constraints of the target node. The model has no representation of the imaging system, of the lithographic process window, or of etch and deposition loading. It has a representation of how layouts look in its training distribution. When the distribution is sparse (and qualified process data is almost always sparse compared to natural images) the model interpolates into regions that visually resemble valid layouts but physically resemble nothing that has ever printed [13], [14], [15]. This is not a problem solved by larger models or more data, because the underlying objective is wrong. A generative model trained to match a distribution of historical layouts will produce historical-looking layouts. It will not produce layouts that satisfy printability and electrical requirements that were never expressed in the training data.

### *Synthetic data that loses causal and physical fidelity*

Synthetic data generation is widely promoted as the solution to the data scarcity that plagues semiconductor AI, where labeled excursions are rare and expensive. [1] Generative adversarial networks and diffusion models are used to produce synthetic wafer-defect maps, synthetic SEM and optical inspection images, synthetic metrology signals, and synthetic fault-detection traces, with the explicit goal of augmenting training sets for downstream defect classifiers and anomaly detectors. [10], [11], [16] The augmented classifiers report improved benchmark accuracy. They also, in production deployment, fail in ways the benchmarks did not predict.

The mechanism is the same as the layout case. A generative model trained to produce visually convincing defect images learns the visual statistics of defects, not the failure physics that produces defects. When the process drifts (a different consumable supplier, a worn chamber part, a tool-to-tool mismatch, a process excursion) the real defect distribution shifts according to the underlying physics. The synthetic distribution, anchored to the training data, does not. The classifier trained on the synthetic-augmented set has learned to recognize the visual hallmarks of the original defect distribution and fails on the shifted real one. The benchmark improvement was real. The production improvement was an illusion that the benchmark could not see.

### *Inverse designs that optimize the learned manifold rather than the physical one*

The most quantitatively measurable failure mode is in inverse design. Latent-space optimization over learned generative priors produces solutions that are optimal with respect to the learned objective and infeasible with respect to physics. A diffusion model trained on optimized photonic or device structures will produce a candidate whose electromagnetic response, when computed by a rigorous Maxwell or TCAD solver, misses the target spectrum or violates the device specification, or whose features fall outside the lithographic process window. The model optimized over a learned manifold that approximated the historical Pareto frontier. The actual Pareto frontier is defined by physics the model never represented. [13], [14], [17] This is not unique to photonics: generative inverse design of semiconductor devices over TCAD-derived manifolds confronts the same non-uniqueness and constraint-satisfaction problem, and recovers physically valid candidates only when the device physics or its constraints are folded into the generator rather than checked after sampling. [18]

Post-hoc filtering, the standard mitigation, addresses the symptom rather than the cause. Each candidate is checked against a physics solver (a lithography simulator, a rigorous EM solver, or a TCAD model), and infeasible candidates are rejected. The cost is twofold. First, the rejection rate is high enough at scale to dominate the compute budget; the savings that motivated generative inverse design in the first place are eroded. Second, and more fundamentally, post-hoc filtering does not improve the model. The next sample from the same prior has the same probability of failure. The model has not learned that the rejected region is infeasible; it has learned nothing, because the filter operates outside the gradient path. [14]

## Architectures for physically constrained generation

The architectural alternatives are no longer hypothetical. Across machine learning research, a coherent toolkit of physics-informed generative methods has emerged over the past three years, and it is well-matched to the problem structure of the fab. The defining feature of these methods is not the addition of a physics-based loss term to an otherwise standard generative model. It is the encoding of physical structure into the architecture itself, such that generated samples satisfy the relevant laws by construction rather than by penalty [19], [20].

**Table 2.** Levels of physics coupling in generative models for constrained physical systems, ordered from weakest to strongest enforcement, with the corresponding role in semiconductor manufacturing. Physics-informed by construction, as used in this perspective, refers specifically to the simulator-in-the-loop and hard-constrained regimes, not to post-hoc screening or to a physics-loss penalty alone.

| Coupling level | Where physics acts, and what it guarantees | Semiconductor example |
|---|---|---|
| Post-hoc screening | Generate first, simulate later. Infeasible candidates are rejected after sampling; the generator itself is unchanged, so the next sample carries the same failure probability. | Generated masks pass through a lithography simulator and are discarded if they fail printability or design-rule checks. |
| Physics-loss regularization | Add PDE-residual or design-rule penalties to the training objective. Physics biases the learned distribution but is not enforced; violations are discouraged, not prevented. | A Maxwell or lithography residual is added to the loss of a photonic-device or mask generator, raising average fidelity without guaranteeing it. |
| Simulator-in-the-loop generation | A differentiable solver guides sampling or optimization, so physics enters the gradient path and steers each step toward feasibility rather than only judging the final output. | Adjoint gradients from a differentiable lithography or electromagnetic solver steer each denoising step of a mask or device sampler. |
| Hard-constrained generation | Outputs are parameterized to satisfy constraints by construction, preventing infeasible candidates. This is the strongest form of physics-informed design. | Layouts are emitted on a design-rule-satisfying parameterization, so every generated candidate is manufacturable by construction. |
| Certified deployment | Independent solver or process qualification remains mandatory before production use, regardless of how the candidate was generated. This layer is orthogonal to the others and never optional. | Masks and recipes still clear independent TCAD, electromagnetic, and process qualification, including reliability flows, before release to a line. |

To make precise what we mean by physics-informed by construction, and to keep the term from being applied too broadly, it is useful to order the available options along a spectrum of how tightly physics enters the generative process (Table 2). These levels are not mutually exclusive; a deployed system typically combines several. They differ sharply, however, in where physics acts and in what they can guarantee.

In this taxonomy, the architectures surveyed below target the simulator-in-the-loop and hard-constrained regimes, which is what we intend by physics-informed by construction. Post-hoc screening and certified deployment remain necessary, but neither is, on its own, what the term denotes.

### *Physics-informed diffusion models*

Diffusion models, including the score-based generative formulation cast as stochastic differential equations [43] can be conditioned on physics-based scoring functions, but the more powerful construction is to embed physical constraints into the denoising trajectory itself. Guidance terms derived from PDE residuals can be added to the score function, steering the reverse diffusion process toward solutions that satisfy the governing equations at each denoising step. The trajectory does not merely end at a physically valid sample; every intermediate sample is biased toward physical validity, which improves both convergence and the quality of the final distribution. For semiconductor applications, this maps onto lithography-constrained mask generation and Maxwell-constrained photonic-device design. [14], [17], [21] Recent semiconductor-relevant demonstrations bear this out: a Maxwell's-equation-based loss embedded directly in the denoising process has been shown to generate photonic and phase-change-material structures with markedly higher physical fidelity than data-only diffusion, [22] and an inverse-lithography physics-informed neural level set enforces the printability and process-window constraints that purely data-driven mask models miss [23].

### *PDE-constrained variational models*

Variational autoencoders can be trained with constraint-aware decoders that map latent codes through a physics solver before producing the final output. The solver becomes part of the model. Gradients flow through the solver during training, which requires that the solver be differentiable, and this differentiability requirement has been the main implementation obstacle. Recent progress on differentiable lithography and imaging solvers, differentiable device (drift-diffusion) models, and differentiable feature-profile and reaction-diffusion process models has substantially expanded the class of fab physics that can be encoded this way. [24] The device physics such decoders must reproduce is itself demanding; rigorous quantum-transport treatments of nanoscale transistors, for instance, capture constitutive behavior that any differentiable surrogate has to match before its gradients can be trusted. [25], [26]

### *Neural-operator priors and structure-preserving generative networks*

Neural operators learn mappings between function spaces and can serve as physics-aware priors in generative pipelines. A neural operator trained to map a mask or layout to its aerial image, or boundary conditions to electromagnetic near fields, can be used as a prior over plausible optical responses, and a generative model layered over this prior produces candidate layouts whose imaging behavior is constrained by what the operator has learned about the lithographic and Maxwell operators. [17], [27], [28] Variational and structure-preserving generative networks encode conservation laws directly into the network parameterization, so that the relevant physical invariants hold along generated trajectories by construction. [29] For the conservative field physics of lithography and on-chip photonics these architectures map cleanly onto Maxwell's equations, while the dissipative transport of etch, deposition, and dopant diffusion is better served by operators that respect continuity and reaction-diffusion structure rather than energy conservation.

Read together (Table 3), these demonstrations trace the gradient that the taxonomy of Table 2 predicts. Where physics enters only as a loss penalty, the reported benefit is higher physical fidelity of individual samples but little change in search cost. Where physics enters the gradient path or the architecture itself, the benefits compound: not only higher validity but order-of-magnitude reductions in simulator calls, because the generator stops proposing infeasible candidates in the first place. The pattern is the central claim of this perspective in miniature. Post-hoc filtering can raise the validity of what survives, but only construction-level coupling converts physical law into search efficiency, bringing the workflow closer to a physics-grounded closed loop that an autonomous fab could run without a human rejecting samples between every step.

**Table 3.** Representative physics-informed generative demonstrations relevant to the fab, organized by where physical structure enters the model and the effect reported relative to data-only or post-hoc-filtered baselines.

| Demonstration (Ref.) | Fab or photonics task | Where physics enters | Reported effect vs. data-only or post-hoc baseline |
|---|---|---|---|
| GAN-OPC [15] | Mask optimization (OPC) | Lithography model as GAN training guidance (loss term) | Generates mask corrections in a forward pass, replacing iterative OPC and improving printed-edge fidelity over unguided GANs. |
| Inverse-lithography neural level set [23] | Mask synthesis | Physics-informed level set enforcing the imaging model (architecture and loss) | Produces masks meeting printability and process-window constraints that purely data-driven mask models violate. |
| MxDiffusion [22] | Inverse photonic metasurface design | Maxwell's-equation residual in the denoising loss (loss term) | Markedly higher physical fidelity of generated structures than data-only diffusion. |
| Physics-guided, fabrication-aware diffusion [32] | Photonic device inverse design | Adjoint sensitivity gradients injected into each denoising step (gradient path) | Reaches target figures of merit with roughly threefold fewer simulations than state-of-the-art optimizers and orders of magnitude fewer than data-only deep learning, with fabrication-aware geometries. |
| TCAD-augmented autoencoder [18] | Semiconductor device variation ID and inverse design | Generative manifold (VAE) that learns device constraints (architecture) | Recovers physically valid device parameters that satisfy constraints by construction, without post-hoc rejection. |

### *Why this is a different design philosophy, not an incremental tweak*

It is tempting to treat physics-informed generation as an enhancement to standard generative architectures: train a normal model, add a physics loss, expect improvement. This framing understates what has changed. The shift from physics-as-penalty to physics-as-structure changes what the model can and cannot represent. A standard diffusion model can in principle approximate any distribution; in practice it approximates the training distribution and fails outside it. (Figure 2) A physics-informed diffusion model can in principle approximate only distributions consistent with the embedded physics; in practice it generalizes to physically valid regions the training distribution never visited. The first class trades coverage for compliance. The second class trades compliance for coverage. For semiconductor manufacturing, the second trade is the right one. The taxonomy in Figure 2 makes this concrete: current data-driven models concentrate their probability mass in the plausible-but-invalid quadrant, whereas the aim of physics-informed architectures is to relocate that mass into the plausible-and-valid quadrant where production-grade design must operate.

## Integration patterns between generative models and scientific simulators

Physics-informed generative AI is not the only structural change required. The data-driven and physics-based cultures of the fab have evolved largely in parallel, with limited shared infrastructure and few people fluent in both. The transition from current practice to physics-informed generative semiconductor manufacturing will be carried by specific integration patterns where the two cultures meet. We identify four that are tractable today (Figure 3) and that, taken together, define the practical research frontier. Figure 3 lays out these four integration patterns, pairing each generative component with the physics-based component that constrains it and indicating where the coupling enters the gradient path, the loss function, or the embedding space.

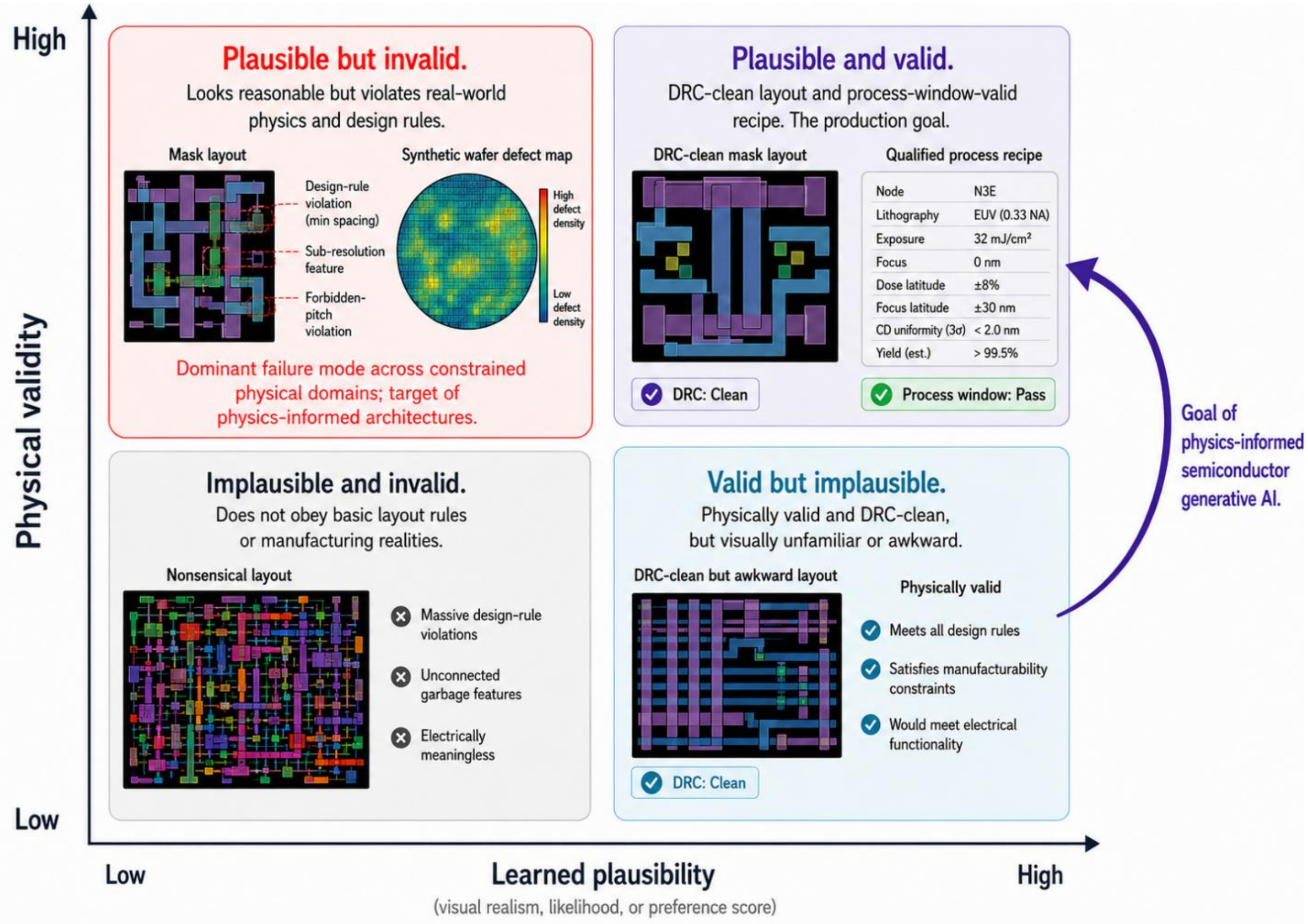

**Figure 2.** Failure mode taxonomy for generative outputs in constrained physical domains, organized along axes of learned plausibility and physical validity. Learned plausibility is whatever the generative objective rewards, whether visual realism, likelihood under the training distribution, or a preference score; in the fab it is typically the visual realism of a layout or defect image. The upper-left quadrant (plausible but invalid) is the dominant failure mode of current data-driven generative models and the principal target of physics-informed architectures. The upper-right quadrant (plausible and valid) is where production-grade generative design in the fab must operate.

### *Constrained generative process planning*

Large language models can reason over fab operations with surprising fluency, drafting process flows, suggesting integration sequences, and explaining process-window allocations. Their failure mode is the same hallucination problem that limits them in code generation: they produce steps the available toolset cannot execute, sequences that violate thermal-budget or contamination constraints, or process parameters that fall outside the qualified process window. The meeting point is to couple the language model to a process-rule and process-window checker that filters and constrains its outputs in the gradient path. The solver provides hard feasibility; the language model provides the search heuristics over the combinatorial space of possible flows. Neither component is sufficient alone. The integrated system is. [12] Domain-adapted large language models for chip design already exhibit this division of labor on tasks such as EDA script generation and engineering question answering, where domain-adaptive pretraining sharply improves reliability over general-purpose models [30].

### *Physics-governed synthetic data generation*

Synthetic data is too valuable to abandon and too dangerous to use without physical grounding. The meeting point is generative models whose sampling distribution is conditioned on a failure-mode physics simulator. Synthetic defects are generated by perturbing litho, etch, and deposition parameters in a physics-based defect-formation model (capturing bridging, voids, particle adders, and CD or overlay errors), then rendering the resulting defects through a learned SEM or optical-inspection appearance model. The visual statistics come from the data; the failure statistics come from the physics. The resulting synthetic distribution shifts correctly when the underlying process drifts, because the physics-based generator tracks the shift. The visual fidelity is high enough to train downstream inspection classifiers. The combination is more useful than either pure data-driven or pure physics-based synthesis [16].

### ***Simulator-conditioned inverse design***

Inverse design with physics-based simulators in the loop has been pursued for years in computational lithography and photonics; [17], [31] the limitation has been the cost of differentiable simulators and the local-minimum behavior of gradient descent over high-dimensional design spaces. Diffusion priors change the picture. Conditioning a diffusion model on simulator gradients produces a generative process that samples from the physically feasible region of design space directly, exploring it broadly rather than descending into the nearest local minimum. The simulator provides the feasibility constraint; the diffusion prior provides the global exploration. Recent demonstrations in inverse lithography, photonic and electromagnetic device design, and process-parameter search show order-of-magnitude reductions in simulator calls per converged design. [14], [15], [31] A concrete instance injects adjoint sensitivity gradients into each denoising step of a diffusion sampler, reaching a target figure of merit with roughly threefold fewer simulations than state-of-the-art nonlinear optimizers and orders of magnitude fewer than purely data-driven deep-learning approaches, while keeping the generated geometries fabrication-aware [32]. Framed in the vocabulary of autonomous experimentation, the diffusion prior acts as a sample-efficient search policy and the differentiable simulator as the reward model, making the loop a physics-grounded counterpart to the Bayesian-optimization and reinforcement-learning strategies used elsewhere for closed-loop design. [39], [40], [41]

### ***Multimodal foundation models trained on text, layout, and simulation***

The largest meeting point, and the most distant, is a foundation model pretrained jointly on natural language, layout and netlist representations, and lithography and TCAD simulation outputs. Such a model would learn shared embeddings across the modalities of the fab rather than over the modalities of the internet. The training infrastructure does not yet exist; the corpora are fragmented and proprietary; the tokenization across modalities is an open problem. The capability, if achieved, would be qualitatively different from anything in current generative semiconductor design. A specification expressed in language could be embedded into the same space as a layout and a simulation result, enabling reasoning that currently requires expert translation at every step. [12] The nearest existing precedent is domain-adaptive pretraining of language models on proprietary chip-design corpora, which already yields large reliability gains within a single modality; extending this to a jointly embedded text, layout, and simulation space is the open frontier. [30]

## Challenges

The architectural shift described above is not free, and several obstacles between current practice and physics-informed generative semiconductor manufacturing deserve direct treatment.

Differentiability of legacy simulators is the most immediate obstacle. The simulation tools that encode the physics of the fab (commercial TCAD suites, proprietary lithography and OPC engines, in-house process and multiphysics codes) are not differentiable. Wrapping them with surrogate models or finite-difference gradient estimators is feasible but expensive and lossy. The path forward requires either differentiable reimplementations, which are emerging in the open-source ecosystem, or differentiable surrogates trained against the legacy solvers, which inherit the accuracy of the surrogate rather than the simulator. [24], [27] Machine-learning regression surrogates of manufacturing processes are already accurate within their training envelope, [33] but as data-bound models they reproduce the statistics of the training set rather than the governing equations, which is precisely the gap a differentiable physics solver is meant to close.

Computational cost of training with physics in the loop is real but often overstated. The per-sample cost is higher than for pure data-driven training, but the sample efficiency is also higher because physical constraints prune the search space. The relevant comparison is total compute to a deployment-ready model, not per-step compute. By the deployment-ready measure, physics-informed approaches are typically competitive and increasingly favorable as differentiable simulator performance improves.

Evaluation metrics are an underappreciated obstacle. The metrics that the generative AI community has built (FID, perceptual scores, likelihood-based measures) are not appropriate for semiconductor outputs. A physics-fidelity benchmark would measure the fraction of generated samples that pass an independent physics solver (a lithography simulator, an EM solver, or a TCAD model), the magnitude of design-rule or governing-equation violation, and the distance from generated samples to known feasible, printable designs. Such benchmarks do not yet exist as community standards. Building them is a prerequisite to comparing approaches honestly. [13]

Certification and qualification for production lines is a longer-horizon challenge. A generative model that produces masks or process recipes intended for high-reliability parts must be qualified under the same regimes that govern conventional design tools and process changes. The qualification pathway for stochastic generative systems is unclear for automotive-grade (for example AEC-Q100) and other high-reliability semiconductor contexts, and for the reliability qualification flows that gate volume production. The community working on physics-informed generative methods will need to engage with these qualification owners directly rather than assume the problem will solve itself.

**Figure 3.** Four integration patterns between the data-driven and physics-based cultures of the fab. Each pattern combines a generative component (language model, generative prior, diffusion sampler, foundation model) with a physics-based component (process-rule and process-window checker, failure-mode simulator, differentiable lithography or TCAD simulator, simulation corpus). The integration is structural rather than post-hoc: the physics enters the gradient path, the loss function, or the embedding space.

## A computational-science research agenda

If the argument of this perspective is accepted, three research priorities follow on illustrative, rather than predictive, horizons. Figure 4 arranges these priorities on a three-horizon timeline, moving from near-term physics-fidelity benchmarks and reference solvers, through medium-term differentiable simulator infrastructure, to the long-term multimodal foundation model.

In the near term (now to 2027), the community should establish physics-fidelity benchmarks for the principal semiconductor generative tasks (layout and mask generation, process-recipe planning, synthetic defect and metrology data, device and mask inverse design). These benchmarks should report not aggregate likelihoods or visual quality but the rate at which generated outputs satisfy independently checked physical and design-rule constraints. They should be open, versioned, and accompanied by reference lithography, EM, and TCAD solvers. Without such benchmarks, the field will continue to optimize for the wrong objective. [13], [14] A partial precedent already exists in computational lithography, where a public benchmark of layout tiles with lithography-model ground truth supports systematic comparison of learned simulation and mask-optimization models; the remaining task is to generalize this benchmarking discipline to physics fidelity across every generative fab task. [34]

In the medium term (2027 to 2030), the infrastructure of differentiable fab simulators should be built out, both as open-source reimplementations of standard lithography, device, and process solvers and as high-fidelity surrogates trained against legacy commercial tools. This is unglamorous engineering work but it is the bottleneck on simulator-in-the-loop generative methods. The investment is comparable to what the community has put into automatic differentiation frameworks for deep learning, and the payoff is comparable. [24]

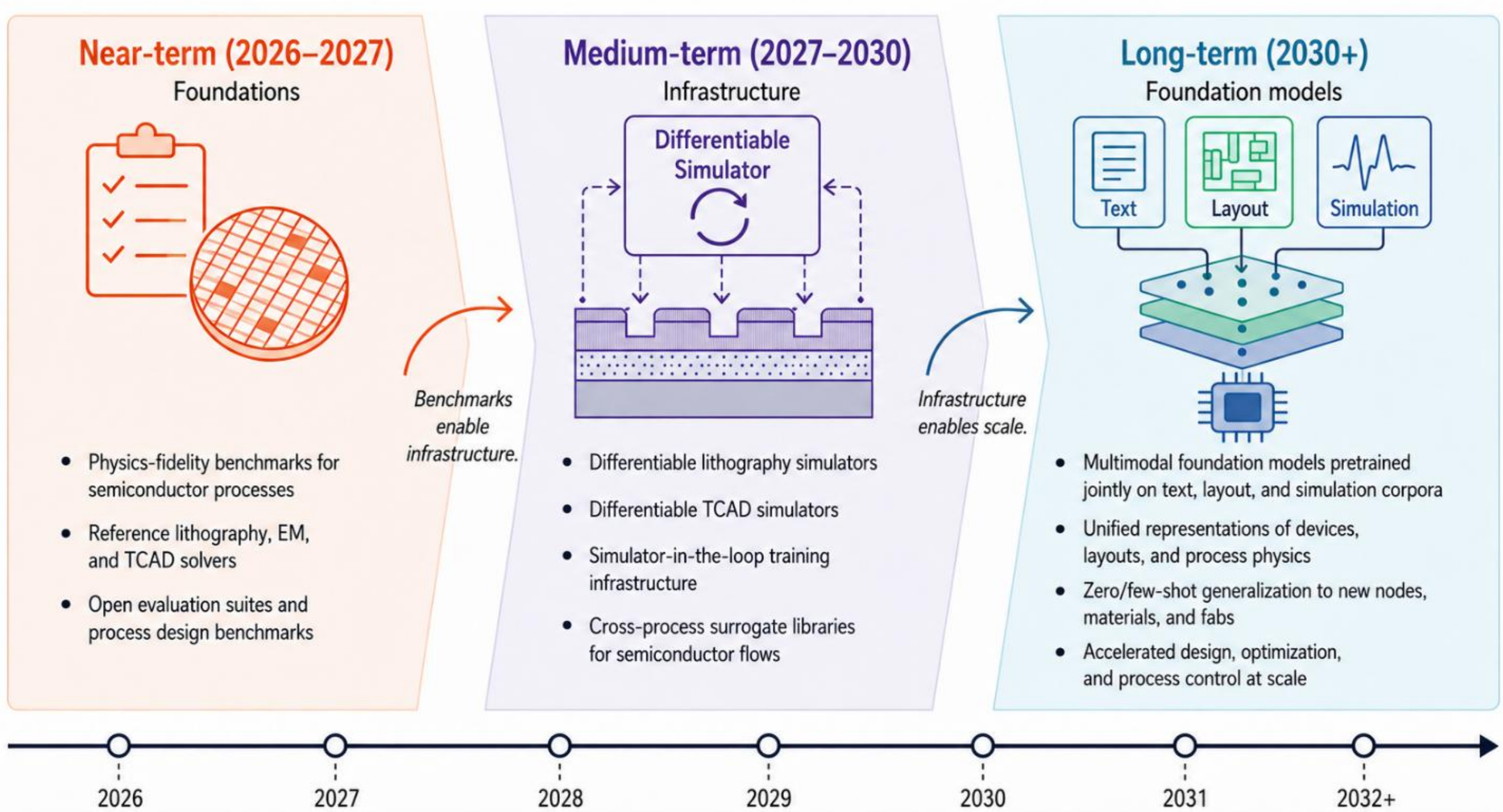

**Figure 4.** Illustrative three-horizon research roadmap for physics-informed generative AI in semiconductor manufacturing. The dates are indicative horizons, not predictions. Near-term (2026 to 2027): physics-fidelity benchmarks and reference solvers. Medium-term (2027 to 2030): differentiable lithography and TCAD simulators and simulator-in-the-loop training infrastructure. Long-term (2030 and beyond): multimodal foundation models pretrained jointly on text, layout, and simulation corpora.

In the long term (2030 and beyond), the foundation-model integration described above (jointly pretrained on text, layout, and simulation) becomes the prize. This will require corpora, compute, and architectural innovation at a scale that no single laboratory or fab will provide. The likely structure is consortium pretraining with federated data contributions, governed by standards bodies that emerge from current industrial AI and semiconductor consortia. The technical work is real, but the organizational work is harder.

## Conclusion

Generative AI is the most consequential capability shift in computational tools since automatic differentiation, and it is arriving in the fab now. The question that matters is not whether generative AI will reshape semiconductor manufacturing; it will. The question is whether the architectures the field adopts will treat physical law as a structural property of the model or as an after-the-fact filter on its outputs. The first choice produces tools that process and integration engineers can trust. The second produces tools that appear compelling in demonstration settings but fail under physical validation.

The choice is not abstract. It is being made now, in every paper, codebase, and product roadmap that imports a generative architecture from language modeling into a semiconductor context. The architectures that respect the electromagnetics, the device physics, and the process constraints of their target domain will define the next decade of advanced semiconductor manufacturing. The architectures that do not will remain confined to early-stage ideation, useful for exploration but not for production. The distinction is architectural: physics is either built into the generator, where it shapes every sample, or bolted on as a filter that can only reject what the model has already produced. Only the first is physics-informed by construction; the second is post-hoc filtering wearing the same name. For semiconductor manufacturing, and for the broader class of constrained physical domains it represents, physical correctness functions less as a constraint on generative AI than as the governing criterion for deployment.

## References

[1] J. Wang, Y. Ma, L. Zhang, R. X. Gao, and D. Wu, “Deep learning for smart manufacturing: Methods and applications,” Journal of Manufacturing Systems, vol. 48, pp. 144–156, 2018.

[2] G. S. May and C. J. Spanos, Fundamentals of Semiconductor Manufacturing and Process Control. Hoboken, NJ, USA: Wiley-IEEE Press, 2006.

[3] J. Moyne and J. Iskandar, “Big data analytics for smart manufacturing: Case studies in semiconductor manufacturing,” Processes, vol. 5, no. 3, p. 39, 2017.

[4] Y. Banadaki, N. Razaviarab, H. Fekrmandi, G. Li, P. Mensah, S. Bai, and S. Sharifi, “Automated quality and process control for additive manufacturing using deep convolutional neural networks,” Recent Progress in Materials, vol. 4, no. 1, 2021.

[5] N. Razaviarab, S. Sharifi, and Y. M. Banadaki, “Smart additive manufacturing empowered by a closed-loop machine learning algorithm,” in Nano-, Bio-, Info-Tech Sensors and 3D Systems III, vol. 10969, pp. 50–57, SPIE, 2019.

[6] Y. Banadaki, “On the use of machine learning for additive manufacturing technology in Industry 4.0,” Journal of Computer Science and Information Technology, vol. 7, pp. 61–68, 2019.

[7] G. Datta, S. S. Sharif, and Y. M. Banad, “Toward digital twins in 3D IC packaging: A critical review of physics, data, and hybrid architectures,” arXiv:2601.23226, 2026.

[8] J. Moyne, E. Del Castillo, and A. M. Hurwitz, Eds., Run-to-Run Control in Semiconductor Manufacturing. Boca Raton, FL, USA: CRC Press, 2001.

[9] H. Kim, J. Y. Hwang, S. E. Kim, Y. C. Joo, and H. Jang, “Thermomechanical challenges of 2.5-D packaging: A review of warpage and interconnect reliability,” IEEE Transactions on Components, Packaging and Manufacturing Technology, vol. 13, no. 10, pp. 1624–1641, 2023.

[10] I. J. Goodfellow, J. Pouget-Abadie, M. Mirza, B. Xu, D. Warde-Farley, S. Ozair, A. Courville, and Y. Bengio, "Generative adversarial nets," in Advances in Neural Information Processing Systems 27 (NeurIPS), 2014, pp. 2672–2680.

[11] J. Ho, A. Jain, and P. Abbeel, "Denoising diffusion probabilistic models," in Advances in Neural Information Processing Systems 33 (NeurIPS), 2020.

[12] L. Makatura, M. Foshey, B. Wang, F. Hähnlein, P. Ma, B. Deng, M. Tjandrasuwita, A. Spielberg, C. E. Owens, P. Y. Chen, A. Zhao, A. Zhu, W. J. Norton, E. Gu, J. Jacob, Y. Li, A. Schulz, and W. Matusik, "How can large language models help humans in design and manufacturing?" arXiv:2307.14377, 2023.

[13] L. Regenwetter, A. Heyrani Nobari, and F. Ahmed, "Deep generative models in engineering design: A review," Journal of Mechanical Design, vol. 144, no. 7, p. 071704, 2022.

[14] F. Mazé and F. Ahmed, "Diffusion models beat GANs on topology optimization," in Proceedings of the AAAI Conference on Artificial Intelligence, vol. 37, no. 8, 2023, pp. 9108–9116.

[15] H. Yang, S. Li, Y. Ma, B. Yu, and E. F. Y. Young, "GAN-OPC: Mask optimization with lithography-guided generative adversarial nets," in Proceedings of the 55th Annual Design Automation Conference (DAC), 2018.

[16] T. Nakazawa and D. V. Kulkarni, "Wafer map defect pattern classification and image retrieval using convolutional neural network," IEEE Transactions on Semiconductor Manufacturing, vol. 31, no. 2, pp. 309–314, 2018.

[17] W. Ma, Z. Liu, Z. A. Kudyshev, A. Boltasseva, W. Cai, and Y. Liu, "Deep learning for the design of photonic structures," Nature Photonics, vol. 15, no. 2, pp. 77–90, 2021.

[18] K. Mehta, S. S. Raju, M. Xiao, B. Wang, Y. Zhang, and H. Y. Wong, "Improvement of TCAD augmented machine learning using autoencoder for semiconductor variation identification and inverse design," IEEE Access, vol. 8, pp. 143519–143529, 2020.

[19] G. E. Karniadakis, I. G. Kevrekidis, L. Lu, P. Perdikaris, S. Wang, and L. Yang, "Physics-informed machine learning," Nature Reviews Physics, vol. 3, no. 6, pp. 422–440, 2021.

[20] M. Raissi, P. Perdikaris, and G. E. Karniadakis, "Physics-informed neural networks: A deep learning framework for solving forward and inverse problems involving nonlinear partial differential equations," Journal of Computational Physics, vol. 378, pp. 686–707, 2019.

[21] D. Shu, Z. Li, and A. Barati Farimani, "A physics-informed diffusion model for high-fidelity flow field reconstruction," Journal of Computational Physics, vol. 478, p. 111972, 2023.

[22] S. Mondal, T. Park, S. Biswas, A. X. Wang, and W. Cai, "MxDiffusion: A physics-aware Maxwell's law-guided diffusion model strategy for inverse photonic metasurface design," Nano Letters, vol. 26, no. 14, pp. 4897–4905, 2026.

[23] X.-Y. Ma and S. Hao, "Inverse lithography physics-informed deep neural level set for mask optimization," Applied Optics, vol. 62, no. 33, pp. 8769–8779, 2023.

[24] Y. Hu, L. Anderson, T.-M. Li, Q. Sun, N. Carr, J. Ragan-Kelley, and F. Durand, "DiffTaichi: Differentiable programming for physical simulation," in International Conference on Learning Representations (ICLR), 2020.

[25] Y. M. M. Banadaki, "Physical modeling of graphene nanoribbon field effect transistor using non-equilibrium Green function approach for integrated circuit design," Ph.D. dissertation, Louisiana State University, 2016.

[26] Y. M. Banadaki and A. Srivastava, "A novel graphene nanoribbon field effect transistor for integrated circuit design," in 2013 IEEE 56th International Midwest Symposium on Circuits and Systems (MWSCAS), pp. 924–927, IEEE, 2013.

[27] Z. Li, N. Kovachki, K. Azizzadenesheli, B. Liu, K. Bhattacharya, A. Stuart, and A. Anandkumar, "Fourier neural operator for parametric partial differential equations," in International Conference on Learning Representations (ICLR), 2021.

[28] L. Lu, P. Jin, G. Pang, Z. Zhang, and G. E. Karniadakis, "Learning nonlinear operators via DeepONet based on the universal approximation theorem of operators," Nature Machine Intelligence, vol. 3, no. 3, pp. 218–229, 2021.

[29] S. Greydanus, M. Dzamba, and J. Yosinski, "Hamiltonian neural networks," in Advances in Neural Information Processing Systems 32 (NeurIPS), 2019.

[30] M. Liu, T.-D. Ene, R. Kirby, C. Cheng, N. Pinckney, R. Liang, et al., "ChipNeMo: Domain-adapted LLMs for chip design," arXiv:2311.00176, 2023.

[31] S. Molesky, Z. Lin, A. Y. Piggott, W. Jin, J. Vučković, and A. W. Rodriguez, "Inverse design in nanophotonics," Nature Photonics, vol. 12, no. 11, pp. 659–670, 2018.

[32] D. Seo, S. Um, S. Lee, J. C. Ye, and H. Chung, "Physics-guided and fabrication-aware inverse design of photonic devices using diffusion models," ACS Photonics, 2026, doi: 10.1021/acsphotonics.5c00993.

[33] M. A. Ilani and Y. M. Banad, "EDMNet: Unveiling the power of machine learning in regression modeling of powder mixed-EDM," The International Journal of Advanced Manufacturing Technology, vol. 135, no. 5, pp. 2555–2570, 2024.

[34] S. Zheng, H. Yang, B. Zhu, B. Yu, and M. D. F. Wong, "LithoBench: Benchmarking AI computational lithography for semiconductor manufacturing," in Advances in Neural Information Processing Systems 36 (NeurIPS), Datasets and Benchmarks Track, 2023.

[35] P. M. Attia, A. Grover, N. Jin, K. A. Severson, et al., "Closed-loop optimization of fast-charging protocols for batteries with machine learning," Nature, vol. 578, no. 7795, pp. 397–402, 2020.

[36] N. J. Szymanski, B. Rendy, Y. Fei, R. E. Kumar, et al., "An autonomous laboratory for the accelerated synthesis of novel materials," Nature, vol. 624, no. 7990, pp. 86–91, 2023.

[37] A. G. Kusne, H. Yu, C. Wu, H. Zhang, et al., "On-the-fly closed-loop materials discovery via Bayesian active learning," Nature Communications, vol. 11, p. 5966, 2020.

[38] E. Stach, B. DeCost, A. G. Kusne, J. Hattrick-Simpers, et al., "Autonomous experimentation systems for materials development: A community perspective," Matter, vol. 4, no. 9, p. 2702, 2021.

[39] P. I. Frazier, "A tutorial on Bayesian optimization," arXiv:1807.02811, 2018.

[40] B. Shahriari, K. Swersky, Z. Wang, R. P. Adams, and N. de Freitas, "Taking the human out of the loop: A review of Bayesian optimization," Proceedings of the IEEE, vol. 104, no. 1, pp. 148–175, 2016.

[41] R. S. Sutton and A. G. Barto, Reinforcement Learning: An Introduction, 2nd ed. Cambridge, MA, USA: MIT Press, 2018.

[42] D. Jones, C. Snider, A. Nassehi, J. Yon, and B. Hicks, "Characterising the digital twin: A systematic literature review," CIRP Journal of Manufacturing Science and Technology, vol. 29, pp. 36–52, 2020.

[43] Y. Song, J. Sohl-Dickstein, D. P. Kingma, A. Kumar, S. Ermon, and B. Poole, "Score-based generative modeling through stochastic differential equations," in International Conference on Learning Representations (ICLR), 2021.